\documentclass[11pt]{article}

\usepackage[preprint]{acl}
\usepackage{times}
\usepackage{latexsym}
\usepackage[T1]{fontenc}
\usepackage[utf8]{inputenc}
\usepackage{microtype}
\usepackage{inconsolata}
\usepackage{graphicx}
\usepackage{booktabs}
\usepackage{amsmath}
\usepackage{amssymb}
\usepackage{colortbl}
\usepackage{pifont}
\usepackage{listings}
\usepackage{stfloats}
\usepackage{placeins}
\usepackage{needspace}

\newcommand{\promptheading}[1]{\Needspace{4\baselineskip}\subsubsection*{#1}}

\definecolor{pubgrey}{gray}{0.42}
\definecolor{rowgrey}{gray}{0.92}

\newcommand{\stackgain}[2]{\begin{tabular}[t]{@{}l@{}}#1\\[-2.5pt]{\tiny(#2)}\end{tabular}}
\newcommand{\diff}[1]{{\tiny\,#1}}
\newcommand{\yes}{\ding{51}}
\newcommand{\no}{\ding{55}}
\newcommand{\sysname}{SupportChain}

\title{\sysname: Trace-Calibrated Aggregation of Quote-Grounded Reasoning Chains for Multi-Hop RAG with Small Readers}

\author{Zhexi Gu\\
  School of Information and Library Science\\
  University of North Carolina at Chapel Hill\\
  \href{mailto:zhexigu@unc.edu}{\texttt{zhexigu@unc.edu}}}
\hypersetup{pdfauthor={Zhexi Gu}}

\begin{document}
\maketitle

\begin{abstract}
Strong results in multi-hop retrieval-augmented question answering come from graph indexes read by models of 20B to 70B parameters or by proprietary models, and building such an index calls a large language model over the whole corpus.
We present \sysname{}, which splits a question into hops, binds each hop's answer to a sentence quoted from a retrieved passage, estimates how reliable each execution is from its own trace, and merges several executions by the support of their answers, the pooled reliability of the executions that agree, stopping once support is high enough.
With a 9B reader under the HippoRAG~2 protocol, \sysname{} reaches 63.0 F1 on MuSiQue, above the best earlier result of 59.2, which used a 20B reader.
With the same reader and embedder, it is significantly better than HippoRAG~2 on all five datasets, by up to 17.2 F1, and it builds no index.
Merging by support is significantly better than majority voting, and than letting the reader choose among the candidates, on three of the five datasets and level with both on the other two. The reliability model needs only 100 answered training questions.
Code is available in a \href{https://anonymous.4open.science/r/KG_reasoning_agent-C69B/}{public anonymous repository}.
\end{abstract}

%======================================================================
\section{Introduction}

Retrieval-augmented generation answers a question from passages fetched from a corpus \citep{lewis2020rag}.
For multi-hop questions, whose answer needs facts from several passages, most recent work adds structure to retrieval: knowledge graphs built by open information extraction \citep{gutierrez2024hipporag,gutierrez2025hipporag2}, a trained graph foundation model \citep{luo2025gfmrag}, or hypergraphs over entities and facts \citep{wang2025hgrag,nguyen2026hyperprove,liu2026hyperaddress}.
The readers behind the strongest of these results have 20B to 70B parameters or are proprietary models.

\begin{figure}[t]
\centering
\includegraphics[width=\columnwidth]{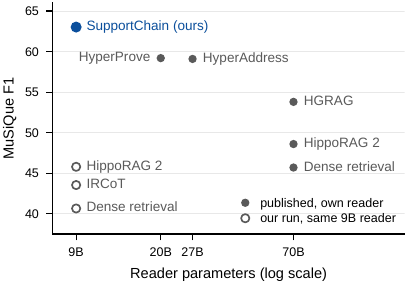}
\caption{MuSiQue test F1 against reader size under the HippoRAG~2 protocol (1,000 questions). Filled grey: published systems with their own readers. Open: baselines we rerun with our 9B reader and the same embedder. Systems read by GPT-4o-mini are only in Table~\ref{tab:main}, since its size is not public.}
\label{fig:size}
\end{figure}

On MuSiQue, the hardest of the standard multi-hop benchmarks, a 9B reader is far behind these results.
With Qwen3.5-9B, HippoRAG~2 reaches 45.8 F1, which is 5.1 above a dense read of the top five passages by the same reader and 13.4 below the best published system.
Its index also costs an LLM pass over the corpus, 8.3M to 15.6M tokens for each corpus of about ten thousand passages.

A 9B reader fails mostly at binding, the step that turns a retrieved passage into the answer of an intermediate hop.
When we decompose MuSiQue development questions and retrieve for each hop, the supporting passage of the first failing hop is retrieved but bound to a wrong answer in 33.2\% of the questions and is missing from retrieval in 5.2\% (Appendix~\ref{app:retrieval}).
A small reader therefore needs evidence it can check at every hop, and a way to tell which of several executions to trust.
\sysname{} has three parts.
A \emph{quote-bound chain} plans the hops, retrieves for each hop, and binds each hop's answer to a sentence quoted from a retrieved passage.
A \emph{reliability model}, a logistic regression over 16 features of the execution trace, predicts whether an execution is right without calling an LLM.
\emph{Support merging} pools the reliabilities of the executions that reach the same answer and returns the answer with the highest support; once support is high, no further executions are run.

Under the HippoRAG~2 protocol, \sysname{} with a 9B reader reaches 63.0 F1 on MuSiQue, above the best published 59.2 obtained with a 20B reader, and averages 73.2 F1 over the three multi-hop datasets, above the best published average of 71.9 (Figure~\ref{fig:size}).
With the same reader and embedder, it beats HippoRAG~2 by 17.2, 6.4, 3.2, 3.0 and 1.6 F1 on MuSiQue, 2WikiMultiHopQA, HotpotQA, NQ and PopQA, significantly on all five.
It spends no LLM tokens on indexing, and choosing the number of executions by support halves the calls on MuSiQue at the same F1.
Support merging beats majority voting by 3.4, 1.8 and 2.2 F1 and letting the reader pick the answer by 3.2, 1.6 and 2.9 F1 on MuSiQue, 2WikiMultiHopQA and NQ, significantly in each case, and is level with both on HotpotQA and PopQA.

Our contributions are:
\begin{itemize}\setlength{\itemsep}{1pt}\setlength{\parskip}{0pt}
\item \textbf{Method.} A quote-bound reasoning chain, a reliability model computed from the execution trace, and support merging with executions added on demand. It builds no graph and trains no LLM.
\item \textbf{Results.} The best MuSiQue score and multi-hop average under the HippoRAG~2 protocol, with a 9B reader, and significant gains over every same-reader baseline on all five datasets.
\item \textbf{Analysis.} Support merging is significantly better than majority voting and than the reader's own choice on three of the five datasets, and never significantly worse than either or than taking the most reliable execution. We show which chain design choices matter and that more complex ones bring no gain.
\end{itemize}

%======================================================================
\section{Related Work}

\paragraph{Structured retrieval.}
GraphRAG \citep{edge2024graphrag}, LightRAG \citep{guo2024lightrag} and RAPTOR \citep{sarthi2024raptor} organise a corpus into graphs or trees before answering.
HippoRAG and HippoRAG~2 extract a knowledge graph with an LLM and run personalised PageRank over it \citep{gutierrez2024hipporag,gutierrez2025hipporag2}; GFM-RAG trains a graph foundation model as retriever \citep{luo2025gfmrag}; HGRAG, HyperProve and HyperAddress build hypergraphs over entities or facts \citep{wang2025hgrag,nguyen2026hyperprove,liu2026hyperaddress}.
All of them call an LLM over the whole corpus before the first question.
\sysname{} builds no index: each hop uses plain dense retrieval.

\paragraph{Iterative retrieval and decomposition.}
IRCoT interleaves retrieval with chain-of-thought steps \citep{trivedi2023ircot}; Self-Ask, ReAct and least-to-most prompting decompose a question into sub-questions \citep{press2023selfask,yao2023react,zhou2023leasttomost}; FLARE, Self-RAG and Adaptive-RAG decide when to retrieve \citep{jiang2023flare,asai2024selfrag,jeong2024adaptiverag}.
Work on attribution asks models to cite their sources \citep{gao2023alce,rashkin2023attribution}.
In \sysname{} every hop returns its answer with a verbatim quote, and the trace an execution leaves is scored without an LLM.

\paragraph{Merging several reasoning paths.}
Self-consistency takes a majority vote over sampled chains \citep{wang2023selfconsistency}; verifiers and step-aware verifiers score chains with trained models \citep{cobbe2021verifiers,li2023diverse,lightman2024verify}; universal self-consistency and multi-chain reasoning let the LLM read several chains and choose \citep{chen2023usc,yoran2023mcr}.
We merge executions of a retrieval chain, take reliability from each execution's own trace, and compare directly against letting the reader choose.
Table~\ref{tab:properties} summarises the differences.

\begin{table}[t]
\centering
\scriptsize
\setlength{\tabcolsep}{2.2pt}
\begin{tabular}{@{}lccccc@{}}
\toprule
 & No LLM & No & Checkable & Calibrated & Cost set \\
Method & at index & training & evidence & reliability & per question \\
\midrule
HippoRAG 2 & \no & \yes & \no & \no & \no \\
GFM-RAG & \no & \no & \no & \no & \no \\
IRCoT & \yes & \yes & \no & \no & \no \\
Self-consistency & \yes & \yes & \no & \no & \no \\
\sysname{} & \yes & \yes & \yes & \yes & \yes \\
\bottomrule
\end{tabular}
\caption{Properties of related methods. \emph{No training}: no retriever or LLM is trained. \emph{Checkable evidence}: each hop returns a quote that can be matched against its passage. \emph{Calibrated reliability}: each execution carries a calibrated estimate of being right (\S\ref{sec:an-reliability}).}
\label{tab:properties}
\end{table}

%======================================================================
\section{Method}

\begin{figure*}[t]
\centering
\includegraphics[width=\textwidth]{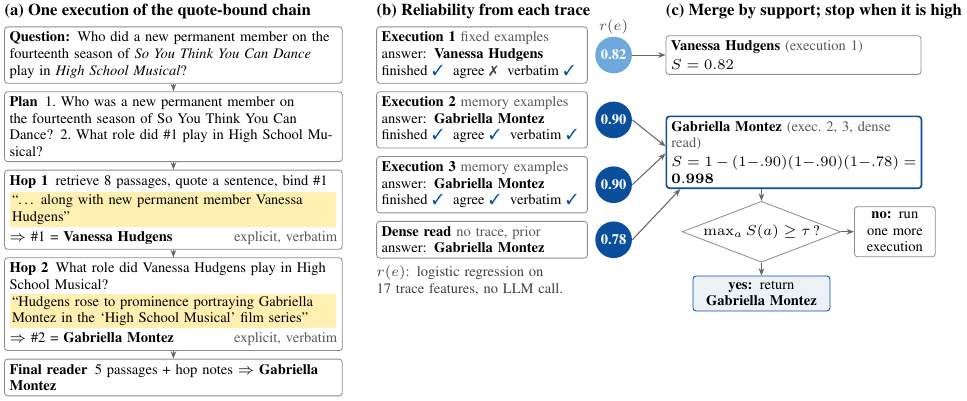}
\caption{\sysname{} on a HotpotQA test question. (a) One execution (execution~2 in b): the plan has two hops; each hop retrieves passages, quotes one sentence (highlighted) and binds an answer; the final reader sees passages and hop notes. (b) Each execution gets a reliability $r(e)$ from features of its trace. Execution~1 binds the role question to a film title while its final reader answers with the actress, so its chain and its reader disagree. (c) Executions with the same answer are pooled into a support $S$; the answer with the highest support is returned, and no more executions run once it passes a threshold.}
\label{fig:overview}
\end{figure*}

\subsection{Problem and Overview}
\label{sec:overview}
Given a question $q$, a passage corpus $\mathcal{P}$, a reader LLM $M$ and an embedder, we want an answer $\hat a$.
An \emph{execution} $e$ of the chain produces an answer $a(e)$ and a trace $\tau(e)$: the plan, the passages retrieved at each hop, the quote and answer bound at each hop, and the final reader's output.
A reliability model maps the trace to $r(e)=\sigma(\mathbf{w}^\top\phi(\tau(e)))\in(0,1)$.
Executions with the same answer form a group $G(a)$, and the answer with the highest support $S(a)$ is returned.
Figure~\ref{fig:overview} shows the three steps on one question: the execution in (a) binds the right role, an execution with a different plan binds a film title instead, and support (c) prefers the answer reached by the two reliable executions and the dense read.

\subsection{Quote-Bound Chain}
\label{sec:chain}
\paragraph{Plan.} The reader splits $q$ into the shortest sequence of single-hop sub-questions, referring to earlier answers as \#1, \#2, \dots, using six fixed examples (four more for comparison questions).
\paragraph{Hop.} Earlier answers are substituted into the sub-question, and the embedder retrieves eight passages from $\mathcal{P}$.
The reader returns the number of the passage it uses, one sentence copied from it, whether the sentence states the answer (\emph{explicit}) or only suggests it (\emph{implied}), and the shortest answer.
If the reader finds no related passage, a second, lenient prompt asks for the most likely answer the passages suggest, again with a quoted sentence.
\paragraph{Final reader.} The final reader uses HippoRAG~2's reading prompt unchanged.
It sees five passages gathered by the hops followed by the hop notes (``Step 1: sub-question $\rightarrow$ answer''), with implied hops marked as likely.
\paragraph{Quotes.} 95\% to 99\% of hops quote a sentence that occurs verbatim in the cited passage (Appendix~\ref{app:retrieval}). All prompts are in Appendix~\ref{app:prompts}.

\subsection{Trace Reliability}
\label{sec:reliability}
The reliability model is a logistic regression over 16 standardised features in four groups:
\emph{completion and agreement} (the chain bound an answer at every hop; the last hop's answer equals the final reader's; the reader produced a marked answer),
\emph{quote evidence} (hops whose quote is verbatim, whose support is explicit, and whose answer occurs in the quote),
\emph{retrieval scores} (scores and ranks of the cited passages), and
\emph{plan shape} (number of hops, links between hops, single-hop plans, an answer copied from the question).
The label is whether the execution's F1 on a training question is at least 0.5.
Computing $r(e)$ calls no LLM, 100 training questions suffice to fit the model (\S\ref{sec:an-reliability}), and no LLM parameter changes.
The dense read has no trace and receives a constant reliability, its share of right answers on the training questions.
Feature definitions and weights are in Appendix~\ref{app:features}.

\subsection{Support Merging}
\label{sec:merge}
Answers are grouped by token-level match after the scorer's normalisation.
The support of an answer is the noisy-OR \citep{pearl1988probabilistic} of the reliabilities in its group,
\begin{equation}
S(a) = 1-\prod_{e\in G(a)}\bigl(1-r(e)\bigr),
\end{equation}
and the output is the answer of the most reliable execution in the group with the highest support.
Majority voting counts every execution as one vote.
Support lets one reliable execution outweigh two unreliable ones that agree: two executions with $r=0.3$ give $S=0.51$, below a single execution with $r=0.8$.

\subsection{Several Executions, On Demand}
\label{sec:ondemand}
We call one run of the chain with the fixed plan examples the \emph{single execution}.
Further executions come from two sources.
\emph{Resampling} decodes the chain again at temperature 0.7 or 0.3.
\emph{Plan examples from memory} replace the fixed plan examples with the six nearest training questions whose plan, written by the chain itself, led to a right answer. A second round adds the right plans of the first round, which gives two \emph{plan executions}.
These memories use only training answers, never gold decompositions.
Section~\ref{sec:an-reliability} compares the two sources.
The \emph{full method} merges four executions: the dense read, the single execution, and the two plan executions.

Two rules save calls.
The \emph{on-demand} rule runs the dense read and the single execution first and runs the two plan executions only when those two disagree (Table~\ref{tab:main}).
The \emph{support threshold} rule adds executions one at a time while the highest support is below $\tau$ (Figure~\ref{fig:cost}).

%======================================================================
\section{Experimental Setup}
\label{sec:setup}

\paragraph{Data.} We use the five test subsets released with HippoRAG~2 \citep{gutierrez2025hipporag2}: MuSiQue \citep{trivedi2022musique}, 2WikiMultiHopQA (2Wiki; \citealp{ho2020constructing}), HotpotQA \citep{yang2018hotpotqa}, NQ \citep{kwiatkowski2019natural} and PopQA \citep{mallen2023trust}, each with 1,000 questions and its passage pool (6,119 to 11,656 passages).
The reliability models and plan memories are fitted on 1,500 questions from each multi-hop training set. For NQ and PopQA we fit nothing: both use the HotpotQA model and memories unchanged, which tests transfer to new datasets. Statistics are in Appendix~\ref{app:stats}.

\paragraph{Baselines.} With the same reader and embedder as ours we run a dense read of the top five passages, IRCoT \citep{trivedi2023ircot} on the three multi-hop datasets, and HippoRAG~2 from its official code, which also uses our reader for extraction.
Published results under the same protocol are listed separately with their readers.

\paragraph{Metrics.} We report F1 and EM with HippoRAG~2's answer extraction and scoring. Intervals are 95\% paired bootstrap intervals over questions with 10,000 resamples \citep{berg2012empirical}; a difference is significant when its interval excludes zero.

\paragraph{Implementation.} Qwen3.5-9B \citep{qwen2026qwen35} with thinking turned off plans, binds and reads, served by vLLM \citep{kwon2023vllm}; NV-Embed-v2 \citep{lee2025nvembed} embeds passages and queries; decoding is greedy. No LLM is trained. The chain and the merging rules were set on development questions drawn from the training sets (Appendix~\ref{app:em}); Appendix~\ref{app:impl} lists all settings.

%======================================================================
\section{Main Results}

\begin{table*}[t]
\centering
\footnotesize
\setlength{\tabcolsep}{2.2pt}
\begin{tabular}{@{}llrr@{\hspace{6pt}}lllllll@{}}
\toprule
System & Reader & Index tok. & Calls/q & MuSiQue & 2Wiki & HotpotQA & NQ & PopQA & Multi-hop & All 5 \\
\midrule
\multicolumn{11}{@{}l}{\textit{\color{pubgrey}Published results under the same protocol, other readers}} \\
\color{pubgrey}Dense retrieval & \color{pubgrey}Llama-3.3-70B & \color{pubgrey}-- & \color{pubgrey}-- & \color{pubgrey}45.7 & \color{pubgrey}61.5 & \color{pubgrey}75.3 & \color{pubgrey}61.9 & \color{pubgrey}55.7 & \color{pubgrey}60.8 & \color{pubgrey}60.0 \\
\color{pubgrey}HippoRAG 2 & \color{pubgrey}Llama-3.3-70B & \color{pubgrey}-- & \color{pubgrey}-- & \color{pubgrey}48.6 & \color{pubgrey}71.0 & \color{pubgrey}75.5 & \color{pubgrey}63.3 & \color{pubgrey}56.2 & \color{pubgrey}65.0 & \color{pubgrey}62.9 \\
\color{pubgrey}HGRAG & \color{pubgrey}Llama-3.3-70B & \color{pubgrey}-- & \color{pubgrey}-- & \color{pubgrey}53.8 & \color{pubgrey}78.3 & \color{pubgrey}76.8 & \color{pubgrey}-- & \color{pubgrey}-- & \color{pubgrey}69.6 & \color{pubgrey}-- \\
\color{pubgrey}IRCoT + GFM-RAG$^a$ & \color{pubgrey}GPT-4o-mini & \color{pubgrey}-- & \color{pubgrey}-- & \color{pubgrey}49.2 & \color{pubgrey}80.8 & \color{pubgrey}71.8 & \color{pubgrey}-- & \color{pubgrey}-- & \color{pubgrey}67.3 & \color{pubgrey}-- \\
\color{pubgrey}HyperProve & \color{pubgrey}GPT-OSS-20B & \color{pubgrey}-- & \color{pubgrey}-- & \color{pubgrey}59.2 & \color{pubgrey}80.0 & \color{pubgrey}76.6 & \color{pubgrey}-- & \color{pubgrey}-- & \color{pubgrey}71.9 & \color{pubgrey}-- \\
\color{pubgrey}HyperAddress$^b$ & \color{pubgrey}Qwen3.5-27B & \color{pubgrey}-- & \color{pubgrey}-- & \color{pubgrey}59.1 & \color{pubgrey}75.1 & \color{pubgrey}80.1 & \color{pubgrey}-- & \color{pubgrey}-- & \color{pubgrey}71.4 & \color{pubgrey}-- \\
\midrule
\multicolumn{11}{@{}l}{\textit{Same reader and embedder as ours, our runs}} \\
Dense read (top 5) & Qwen3.5-9B & 0 & 1.0 & 40.7 & 55.4 & 75.0 & 58.0 & 55.7 & 57.0 & 57.0 \\
IRCoT & Qwen3.5-9B & 0 & 2.1--2.3 & 43.5 & 59.7 & 75.6 & -- & -- & 59.6 & -- \\
HippoRAG 2 & Qwen3.5-9B & 8.3--15.6M & 2.0 & 45.8 & 72.3 & 74.7 & 58.4 & 55.8 & 64.3 & 61.4 \\
\midrule
\multicolumn{11}{@{}l}{\textit{SupportChain (ours), no index}} \\
Single execution & Qwen3.5-9B & 0 & 3.5--5.3 & 59.8 & 78.2 & 76.5 & 59.8 & 57.4 & 71.5 & 66.3 \\
Full method & Qwen3.5-9B & 0 & 11.3--16.9 & \stackgain{\textbf{63.0}$^\dagger$}{+17.2} & \stackgain{\underline{78.6}$^\dagger$}{+6.4} & \stackgain{\textbf{78.0}$^\dagger$}{+3.2} & \stackgain{\underline{61.4}$^\dagger$}{+3.0} & \stackgain{57.4$^\dagger$}{+1.6} & \stackgain{\underline{73.2}}{+8.9} & \stackgain{\underline{67.7}}{+6.3} \\
On demand & Qwen3.5-9B & 0 & 5.7--12.5 & \underline{62.9} & 78.2 & \underline{77.8} & \textbf{62.1} & \underline{57.5} & 73.0 & \underline{67.7} \\
Single + 2 samples & Qwen3.5-9B & 0 & 11.2--16.3 & 62.6 & \textbf{79.5} & \textbf{78.0} & \underline{61.4} & \textbf{59.0} & \textbf{73.4} & \textbf{68.1} \\
\bottomrule
\end{tabular}
\caption{Test F1 (1,000 questions per dataset). Published results: dense retrieval and HippoRAG~2 \citep{gutierrez2025hipporag2}, HGRAG \citep{wang2025hgrag}, IRCoT + GFM-RAG \citep{luo2025gfmrag}, HyperProve \citep{nguyen2026hyperprove}, HyperAddress \citep{liu2026hyperaddress}; IRCoT: \citet{trivedi2023ircot}. \emph{Multi-hop} averages MuSiQue, 2Wiki and HotpotQA; \emph{All 5} averages all datasets. \emph{Index tok.}: LLM tokens spent on each corpus before the first question. \emph{Calls/q}: LLM calls per question, range over the datasets. In the two lower groups, the best value of each column is bold and the second is underlined. $^\dagger$: the full method is significantly above the best same-reader baseline (paired bootstrap, 95\%). Small numbers under the full method: gain over HippoRAG~2 with the same reader. $^a$ HotpotQA pool of 9,221 passages. $^b$ reported on the last 800 of the 1,000 questions. \emph{Single + 2 samples}: the single execution merged with two samples at $T{=}0.7$ (\S\ref{sec:an-reliability}). IRCoT was run on the three multi-hop datasets only. EM scores are in Appendix~\ref{app:em}.}
\label{tab:main}
\end{table*}

\subsection{Accuracy}
\label{sec:accuracy}
Table~\ref{tab:main} gives three results.

\paragraph{Same reader.} With Qwen3.5-9B and NV-Embed-v2 for every system, the full method beats HippoRAG~2 by 17.2, 6.4, 3.2, 3.0 and 1.6 F1, significantly on all five datasets. It beats IRCoT by 19.5, 18.9 and 2.4 on the multi-hop datasets and the dense read by 22.4, 23.2, 3.0, 3.4 and 1.7.

\paragraph{One execution is already strong.} The single execution beats the same-reader HippoRAG~2 by 14.0, 5.9, 1.7, 1.4 and 1.6 F1, significantly on MuSiQue, 2Wiki and PopQA. Most of the gap to graph retrieval comes from the quote-bound chain itself; merging adds the rest.

\paragraph{Published results.} On MuSiQue, 63.0 F1 is the highest under this protocol (previously 59.2), and EM is 7 points above the best published EM. The multi-hop average of 73.2 is also the highest (previously 71.9). The best single results on 2Wiki (80.8) and HotpotQA (80.1) are about 2 F1 ahead of ours. They come from GPT-4o-mini with a trained graph retriever and from a 27B reader, and no published system in Table~\ref{tab:main} is ahead of ours on more than one of the three multi-hop datasets.

\subsection{Cost}
\label{sec:cost}

\begin{figure}[t]
\centering
\includegraphics[width=\columnwidth]{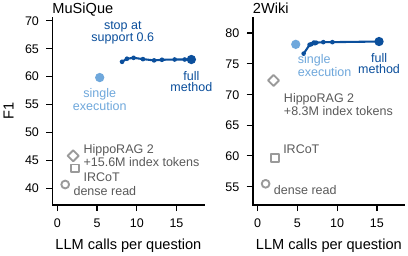}
\caption{Test F1 against LLM calls per question. The line adds plan executions one at a time while the highest support is below a threshold (0.5 to 0.98, left to right, then all executions). HippoRAG~2 also needs an LLM-built index; its token count is printed next to it. Other datasets: Appendix~\ref{app:cost}.}
\label{fig:cost}
\end{figure}

\sysname{} needs no index, while HippoRAG~2 spends 8.3M to 15.6M LLM tokens per corpus, a cost that grows with the corpus (Table~\ref{tab:main}).
Counting the index plus 1,000 questions, the single execution uses a quarter to a half of HippoRAG~2's tokens (MuSiQue 7.1M against 20.8M; 2Wiki 6.8M against 13.4M), and the full method on demand uses about as many or fewer (MuSiQue 17.5M against 20.8M; 2Wiki 13.7M against 13.4M; HotpotQA 9.0M against 18.8M).
On MuSiQue dev-A, a support threshold of 0.6 keeps F1 within 0.4 of all executions (63.4 against 63.8) with half the calls (8.9 against 17.3). With the same threshold on test (Figure~\ref{fig:cost}), the method reaches 63.2 F1 with 8.8 calls per question, against 63.0 with all executions and 16.9 calls, and every threshold from 0.6 to 0.98 stays within 0.3 F1 of the full method.
The reliability model and plan memories are prepared once per dataset from the training questions (Appendix~\ref{app:cost}). This preparation does not depend on the corpus, NQ and PopQA reuse that of HotpotQA, and the single execution needs none.

%======================================================================
\section{Analysis}

\subsection{How to Merge}
\label{sec:an-merge}

\begin{table*}[t]
\centering
\footnotesize
\setlength{\tabcolsep}{2.6pt}
\begin{tabular}{@{}l lllll lllll@{}}
\toprule
& \multicolumn{5}{c}{Full method (dense read + single + 2 plans)} & \multicolumn{5}{c}{Single + 2 samples ($T{=}0.7$)} \\
\cmidrule(lr){2-6}\cmidrule(l){7-11}
Merge rule & MuSiQue & 2Wiki & Hotpot & NQ & PopQA & MuSiQue & 2Wiki & Hotpot & NQ & PopQA \\
\midrule
Majority vote & 59.6\diff{+3.4}$^\dagger$ & 76.9\diff{+1.8}$^\dagger$ & 78.3\diff{$-$0.3} & 59.2\diff{+2.2}$^\dagger$ & 57.5\diff{0.0} & 60.9\diff{+1.8}$^\dagger$ & 78.5\diff{+1.1}$^\dagger$ & 77.2\diff{+0.8}$^\dagger$ & 61.3\diff{+0.1} & 58.0\diff{+1.0}$^\dagger$ \\
Most reliable exec. & 62.7\diff{+0.3} & 78.2\diff{+0.4} & 77.9\diff{+0.1} & 60.4\diff{+1.1} & 55.6\diff{+1.9}$^\dagger$ & 62.0\diff{+0.7} & 79.7\diff{$-$0.1} & 77.7\diff{+0.3} & 59.9\diff{+1.6}$^\dagger$ & 59.1\diff{$-$0.2} \\
Reader picks & 59.8\diff{+3.2}$^\dagger$ & 77.0\diff{+1.6}$^\dagger$ & 77.9\diff{0.0} & 58.5\diff{+2.9}$^\dagger$ & 57.3\diff{+0.2} & -- & -- & -- & -- & -- \\
\midrule
\textbf{Support (ours)} & \textbf{63.0} & \textbf{78.6} & \textbf{78.0} & \textbf{61.4} & \textbf{57.4} & \textbf{62.6} & \textbf{79.5} & \textbf{78.0} & \textbf{61.4} & \textbf{59.0} \\
\midrule
\color{pubgrey}Oracle & \color{pubgrey}69.6 & \color{pubgrey}81.8 & \color{pubgrey}82.7 & \color{pubgrey}69.5 & \color{pubgrey}61.5 & \color{pubgrey}68.9 & \color{pubgrey}82.0 & \color{pubgrey}82.2 & \color{pubgrey}69.2 & \color{pubgrey}63.3 \\
\bottomrule
\end{tabular}
\caption{The same executions merged by different rules (test F1). Small numbers: support minus the rule, paired; $^\dagger$: the paired 95\% interval excludes zero. \emph{Reader picks}: the reader sees every distinct answer with the full traces that reached it and chooses one (one extra call when the executions disagree). \emph{Oracle} picks the best execution using the gold answer.}
\label{tab:merge}
\end{table*}

\begin{figure}[t]
\centering
\includegraphics[width=\columnwidth]{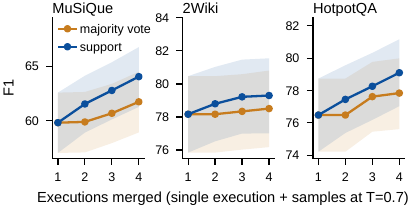}
\caption{Test F1 against the number of executions merged by support or majority vote. The first execution is the single execution; further ones are samples at $T{=}0.7$. Each point averages all subsets of that size; bands are 95\% bootstrap intervals.}
\label{fig:executions}
\end{figure}

Table~\ref{tab:merge} merges the same executions with different rules.
For the executions of the full method, support is significantly better than majority voting on MuSiQue, 2Wiki and NQ (by 3.4, 1.8 and 2.2 F1) and level on HotpotQA and PopQA.
When the single execution is merged with two samples instead (right half), support is significantly better than voting on four datasets and level on NQ (Appendix~\ref{app:sources}).
We also let the reader choose: it sees every distinct answer with the full traces that reached it (each hop's sub-question, answer and quote) and uses one more call.
It reaches 59.8, 77.0, 77.9, 58.5 and 57.3 F1. Support is higher by 3.2, 1.6, 0.0, 2.9 and 0.2, significantly on three datasets, so the reader's choice is about as good as a majority vote.
Taking the most reliable execution without pooling is closer: support is ahead of it on all five datasets, significantly on PopQA (+1.9).
The oracle execution is still 3 to 8 points above support, which leaves room for better selection.
Figure~\ref{fig:executions} adds executions one by one: support and majority vote start from the same point, support stays above, and the gap does not shrink with more executions.

\subsection{The Reliability Model}
\label{sec:an-reliability}

\begin{figure}[t]
\centering
\includegraphics[width=\columnwidth]{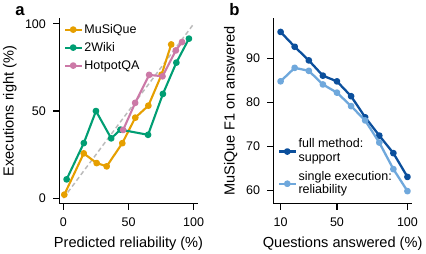}
\caption{(a) Calibration of the reliability of individual chain executions on test (the single execution and the two plan executions, 3,000 per dataset; bins with at least 20). ECE: 7.0 MuSiQue, 6.4 2Wiki, 2.8 HotpotQA. (b) Selective answering on MuSiQue: F1 on the answered questions when only those with the highest support (full method) or reliability (single execution) are answered.}
\label{fig:reliability}
\end{figure}

\begin{table}[t]
\centering
\footnotesize
\setlength{\tabcolsep}{3pt}
\begin{tabular}{@{}lrrr@{}}
\toprule
Reliability model & MuSiQue & 2Wiki & Hotpot \\
\midrule
\multicolumn{4}{@{}l}{\textit{Remove one feature group and refit}} \\
\quad Completion and agreement & $-$1.70$^\dagger$ & $-$0.41$^\dagger$ & $-$0.36 \\
\quad Quote evidence & +0.04 & +0.60 & +0.06 \\
\quad Retrieval scores & $-$0.14 & $-$0.01 & $-$0.13 \\
\quad Plan shape & +0.18 & +0.05 & +0.01 \\
\quad All (constant) & $-$2.16$^\dagger$ & +0.16 & $-$0.61$^\dagger$ \\
\midrule
\multicolumn{4}{@{}l}{\textit{Training questions (full method 1,500)}} \\
\quad 100 & $-$0.49 & +0.21 & $-$0.01 \\
\quad 250 & $-$0.07 & +0.24 & +0.43 \\
\quad 500 & $-$0.59$^\dagger$ & +0.28 & +0.05 \\
\quad 1,000 & $-$0.56$^\dagger$ & +0.04 & 0.00 \\
\quad 3 datasets pooled & $-$0.25 & +0.64 & +0.19 \\
\midrule
\multicolumn{4}{@{}l}{\textit{Fitted on another dataset}} \\
\quad MuSiQue & -- & $-$7.53$^\dagger$ & +0.36 \\
\quad 2Wiki & $-$1.88$^\dagger$ & -- & $-$0.37 \\
\quad HotpotQA & $-$0.58 & +0.74$^\dagger$ & -- \\
\bottomrule
\end{tabular}
\caption{Changes to the reliability model: test F1 of the full method minus that of the full method with its own model (paired). $^\dagger$: the 95\% interval excludes zero.}
\label{tab:reliability}
\end{table}

\paragraph{Calibration.} The reliability of an individual chain execution tracks its accuracy closely, with ECE \citep{guo2017calibration} of 7.0, 6.4 and 2.8 on the multi-hop datasets (Figure~\ref{fig:reliability}a).
\paragraph{Abstention.} Support also ranks questions for selective answering \citep{kamath2020selective}: answering only the half with the highest support gives 84.7, 94.9, 85.2, 76.2 and 69.3 F1 on the five datasets (MuSiQue in Figure~\ref{fig:reliability}b).
\paragraph{Training data.} A model fitted on only 100 training questions is not significantly different from the one used by the full method ($-$0.49, +0.21, $-$0.01; Table~\ref{tab:reliability}), and one model for the three datasets pooled is as good ($-$0.25, +0.64, +0.19).
\paragraph{Features.} The completion and agreement group does most of the work: removing it costs 1.7 F1 on MuSiQue. Removing any other group stays within the intervals.
\paragraph{Where executions come from.} Plan examples from memory and resampling tie on four datasets, and resampling is significantly better on PopQA (59.0 against 57.4; Appendix~\ref{app:sources}). The single execution plus two samples is significantly above the single execution on all five datasets (+2.8, +1.3, +1.5, +1.6, +1.6). The method needs no memory beyond the reliability model; resampling is enough.

\subsection{Chain Design}
\label{sec:an-chain}

\begin{table}[t]
\centering
\footnotesize
\setlength{\tabcolsep}{2pt}
\begin{tabular}{@{}lrrr@{}}
\toprule
Variant & MuSiQue & 2Wiki & Hotpot \\
\midrule
\multicolumn{4}{@{}l}{\textit{Design choices}} \\
\quad Reader sees notes only & $-$4.89$^\dagger$ & $-$2.48$^\dagger$ & $-$8.37$^\dagger$ \\
\quad Reader sees passages only & $-$4.64$^\dagger$ & $-$0.65 & +1.92$^\dagger$ \\
\quad No final reader & $-$1.49$^\dagger$ & $-$45.75$^\dagger$ & $-$21.93$^\dagger$ \\
\quad 4 passages per hop & $-$0.12 & $-$0.52 & +0.54 \\
\midrule
\rowcolor{rowgrey}\multicolumn{4}{@{}l}{\textit{More complex, no gain}} \\
\rowcolor{rowgrey}\quad 3 candidates per hop & $-$2.01$^\dagger$ & $-$0.99 & +0.42 \\
\rowcolor{rowgrey}\quad + scorer, backtracking$^\ddagger$ & $-$0.73 & -- & -- \\
\rowcolor{rowgrey}\quad Gold-decomposition plans$^\ddagger$ & $-$1.56 & -- & -- \\
\bottomrule
\end{tabular}
\caption{One change to the single execution at a time: test F1 of the changed chain minus that of the single execution (paired). \emph{Reader}: the final reader; 4 passages per hop instead of 8. $^\dagger$: the 95\% interval excludes zero. $^\ddagger$ uses gold decompositions of training questions; run on MuSiQue only.}
\label{tab:chain}
\end{table}

Table~\ref{tab:chain} changes one design choice of the single execution at a time.
A final reader that sees both hop notes and passages is the most stable choice: notes alone lose 4.9, 2.5 and 8.4 F1, and dropping the final reader so that the last hop answers loses 45.8 on 2Wiki, whose comparison questions need a final step after the hops. Passages alone help on HotpotQA (+1.9) but lose 4.6 on MuSiQue.
Four passages per hop work about as well as eight.
More complex designs bring no gain: with several candidates per hop, a learned hop scorer with backtracking, or plan examples built from gold decompositions, no chain is significantly better than the single execution, and three candidates per hop are significantly worse on MuSiQue.

\subsection{Where the Errors Are}
\label{sec:an-errors}

\begin{figure}[t]
\centering
\includegraphics[width=\columnwidth]{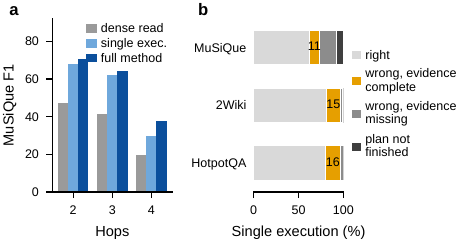}
\caption{(a) MuSiQue test F1 by number of hops (518, 316 and 166 questions). (b) Outcomes of the single execution on test: right; wrong although all gold passages were in front of the reader (\%, printed); wrong with the plan finished but evidence missing; plan not finished.}
\label{fig:errors}
\end{figure}

Retrieval inside the chain is already better than in the baselines: among the five passages given to the reader, gold-evidence recall on MuSiQue is 81.6 for the chain against 69.3, 71.6 and 74.0 for the dense read, IRCoT and HippoRAG~2 (Appendix~\ref{app:retrieval}).
What remains is mostly binding and reading.
On MuSiQue, 27.7\% of the questions have no right execution at all, and this share rises to 52.4\% on 4-hop questions, where F1 is lowest (Figure~\ref{fig:errors}a).
On 2Wiki and HotpotQA, most wrong answers of the single execution had all gold passages in front of the reader (Figure~\ref{fig:errors}b).
\citet{wei2026beyond} observe a similar shift on MuSiQue with GPT-4o-mini: once retrieval gaps are closed, errors after retrieval become a larger share.

\subsection{Case Study}
\label{sec:an-case}

% Case from data/cases.jsonl: MuSiQue test question 3hop1__31995_24918_24939,
% kind "support right where the majority vote is wrong".
\begin{table}[t]
\centering
\scriptsize
\setlength{\tabcolsep}{2pt}
\renewcommand{\arraystretch}{1.08}
\begin{tabular}{@{}p{0.985\columnwidth}@{}}
\toprule
\textbf{Q:} The leader visiting where Steven Spielberg's grandparents are from met with whom on November 22? \quad \textbf{Gold:} Pope John Paul II \\
\midrule
\textbf{Single execution} (fixed plan examples), $r=0.73$ \\
1.~Where are Steven Spielberg's grandparents from? $\rightarrow$ \textbf{Ukraine} (explicit)\newline
\hspace*{1em}\textit{``His paternal grandparents were immigrants from Ukraine who settled in Cincinnati \dots''} \\
2.~Who was the leader visiting Ukraine on November 22? $\rightarrow$ \textbf{Pope John Paul II} (implied)\newline
\hspace*{1em}\textit{``On November 26, 1989, \dots\ on the eve of a meeting between Pope John Paul II and Soviet President Gorbachev.''} \\
3.~Who did Pope John Paul II meet on November 22? $\rightarrow$ \textbf{Soviet President Gorbachev} (implied, same quote) \\
Answer: Soviet President Gorbachev \hfill F1 0.0 \\
\midrule
\textbf{Plan execution from memory} (two identical executions), $r=0.76$ each \\
1.~Where are Steven Spielberg's grandparents from? $\rightarrow$ \textbf{Ukraine} (explicit) \\
2.~Who is the leader visiting Ukraine? $\rightarrow$ \textbf{Gorbachev} (explicit)\newline
\hspace*{1em}\textit{``The demonstrations coincided with a visit to Ukraine by Soviet President Gorbachev.''} \\
3.~Who did Gorbachev meet with on November 22? $\rightarrow$ \textbf{Pope John Paul II} (implied, quote of the meeting) \\
Answer: Pope John Paul II \hfill F1 1.0 \\
\midrule
\textbf{Dense read}, $r=0.46$: a long ``Thought:'' that names both men and gives no final answer \hfill F1 0.01 \\
\midrule
\textbf{Majority vote:} the dense read's text \hfill F1 0.01 \\
\textbf{Support:} Pope John Paul II, $S=0.99$ \hfill F1 1.0 \\
\bottomrule
\end{tabular}
\caption{A MuSiQue test question. The single execution binds the wrong leader at hop~2 from an implied quote; the plan executions bind the visiting leader from an explicit quote. All quotes occur verbatim in the retrieved passages.}
\label{tab:case}
\end{table}

In Table~\ref{tab:case}, the single execution binds the wrong leader at hop~2 from an implied quote about a meeting on another date, and then answers Gorbachev.
The two plan executions bind Gorbachev from an explicit, verbatim quote and reach the gold answer with a higher reliability.
The dense read reasons at length without a final answer; its text names both men, so all four outputs fall into one answer group, and majority voting returns that group's first member, the dense read (Appendix~\ref{app:merge}). Support returns the answer of the group's most reliable execution, Pope John Paul~II.
Appendix~\ref{app:cases} shows more cases, including failures.

%======================================================================
\section{Conclusion}
\sysname{} answers multi-hop questions with a chain that binds every hop to a quoted sentence, scores each execution from its trace, and merges executions by the support pooled from these scores.
With a 9B reader and no index, it sets the best MuSiQue score and multi-hop average under the HippoRAG~2 protocol and beats every same-reader baseline on five datasets.
The errors that remain for a small reader are in binding and reading at each hop, and that is where we expect the next gains.

%======================================================================
\section*{Limitations}
\begin{itemize}\setlength{\itemsep}{1pt}\setlength{\parskip}{0pt}
\item All results use one reader (Qwen3.5-9B) and one embedder (NV-Embed-v2). We have not run the method with larger readers, so the 2 F1 by which the best published systems lead on 2Wiki and HotpotQA cannot yet be split into reader size and method.
\item The corpora are the passage pools fixed by the protocol, about ten thousand passages per dataset. We have not tested full Wikipedia.
\item The reliability model needs about 100 answered questions of the same kind. A model fitted on one dataset does not always carry over to another (the MuSiQue model loses 7.5 F1 on 2Wiki). One model fitted on the three datasets pooled matches the per-dataset models, and the HotpotQA model serves NQ and PopQA.
\item Noisy-OR treats executions of the same reader as independent, so the merged support overstates the probability of a right answer (ECE 15 to 25 on the multi-hop datasets). It still ranks questions well for selective answering (\S\ref{sec:an-reliability}), and the calibrated quantity is the reliability of an individual execution.
\item Answering takes more LLM calls than single-shot retrieval methods. When questions are many and the corpus is small, a method that builds its index once uses fewer tokens in total.
\end{itemize}

\bibliography{references}

%======================================================================
\appendix
\setcounter{topnumber}{4}
\setcounter{bottomnumber}{3}
\setcounter{totalnumber}{6}
\renewcommand{\topfraction}{0.9}
\renewcommand{\bottomfraction}{0.8}
\renewcommand{\textfraction}{0.08}
\renewcommand{\floatpagefraction}{0.75}
\raggedbottom
\setcounter{dbltopnumber}{4}
\renewcommand{\dbltopfraction}{0.9}
\renewcommand{\dblfloatpagefraction}{0.75}
%======================================================================
\section{Development Results and Test EM}
\label{app:em}

\begin{table*}[!htbp]
\centering
\footnotesize
\setlength{\tabcolsep}{6pt}
\begin{tabular}{@{}lrrrrr@{}}
\toprule
System & MuSiQue & 2Wiki & HotpotQA & NQ & PopQA \\
\midrule
Dense retrieval (published) & 34.7 & 57.5 & 62.8 & -- & -- \\
HippoRAG 2 (published) & 37.2 & 65.0 & 62.7 & -- & -- \\
HGRAG (published) & 42.2 & 70.3 & 63.9 & -- & -- \\
IRCoT + GFM-RAG (published) & 36.6 & 72.5 & 56.0 & -- & -- \\
HyperProve (published) & 42.8 & 69.7 & 63.9 & -- & -- \\
HyperAddress (published) & 46.5 & 66.8 & 67.1 & -- & -- \\
\midrule
Dense read, our reader & 32.3 & 50.6 & 62.6 & 44.7 & 42.8 \\
IRCoT, our reader & 34.2 & 54.1 & 63.3 & -- & -- \\
HippoRAG 2, our reader & 37.3 & 64.9 & 62.6 & 45.6 & 43.8 \\
Single execution & 50.3 & 70.5 & 63.9 & 48.2 & 45.1 \\
Full method & 53.6 & 70.8 & 64.8 & 49.6 & 45.0 \\
Full method, on demand & 53.1 & 70.5 & 64.7 & 50.0 & 45.2 \\
Single exec. + 2 samples & 53.3 & 71.8 & 65.1 & 49.8 & 46.6 \\
\bottomrule
\end{tabular}
\caption{Test EM, same systems as Table~\ref{tab:main}. Published EM for NQ and PopQA is not included.}
\label{tab:em}
\end{table*}

\begin{table}[!htbp]
\centering
\footnotesize
\setlength{\tabcolsep}{2.5pt}
\begin{tabular}{@{}lccc@{}}
\toprule
System & MuSiQue & 2Wiki & HotpotQA \\
\midrule
Dense read & 43.7 / 34.6 & 62.1 / 59.6 & 75.1 / 61.0 \\
IRCoT & 45.6 / 36.4 & 68.3 / 64.4 & 75.7 / 61.6 \\
HippoRAG 2 & 44.2 / 35.4 & 84.4 / 79.6 & 79.1 / 64.4 \\
Single execution & 59.2 / 50.8 & 89.2 / 83.4 & 77.6 / 62.8 \\
Full method & 63.8 / 55.2 & 89.0 / 83.4 & 78.6 / 63.0 \\
On demand & 63.1 / 54.2 & 89.2 / 83.6 & 79.2 / 63.8 \\
\bottomrule
\end{tabular}
\caption{Development set dev-A (500 questions per dataset), F1 / EM, reader Qwen3.5-9B.}
\label{tab:dev}
\end{table}

Table~\ref{tab:em} gives test EM for the systems of Table~\ref{tab:main}, and Table~\ref{tab:dev} gives F1 and EM on the development set dev-A.
dev-A was drawn from the training split of each dataset, while the test subsets come from the official development sets, so the two distributions differ.
On MuSiQue the median cosine similarity between a question and its nearest training question is 0.535 on dev-A and 0.291 on test, so plan examples from memory find closer neighbours on dev-A (Appendix~\ref{app:variants}).
On 2Wiki every system scores higher on dev-A than on test (HippoRAG~2: 84.4 against 72.3; the full method: 89.0 against 78.6). For the full method the gap is almost entirely in compositional questions (78.4 against 58.1), where 96\% of test questions have all evidence in front of the reader and most wrong answers are right in content but written differently from the single gold answer (\emph{Filipino} for \emph{Philippines}, \emph{American} for \emph{United States}).

%======================================================================
\section{Datasets}
\label{app:stats}

\begin{table*}[!htbp]
\centering
\footnotesize
\begin{tabular}{@{}lrrrrl@{}}
\toprule
Dataset & Test questions & Passages & Gold passages / q & Words / q & Training questions for reliability and plans \\
\midrule
MuSiQue & 1,000 & 11,656 & 2.65 & 18.0 & 1,500 from MuSiQue \\
2Wiki & 1,000 & 6,119 & 2.47 & 12.0 & 1,500 from 2Wiki \\
HotpotQA & 1,000 & 9,811 & 2.00 & 16.0 & 1,500 from HotpotQA \\
NQ & 1,000 & 9,633 & 3.79 & 9.2 & 1,500 from HotpotQA \\
PopQA & 1,000 & 8,676 & 2.00 & 6.3 & 1,500 from HotpotQA \\
\bottomrule
\end{tabular}
\caption{Datasets. Test question types: MuSiQue 2 hops 518, 3 hops 316, 4 hops 166; 2Wiki bridge comparison 235, comparison 244, compositional 413, inference 108; HotpotQA bridge 811, comparison 189. Nothing is fitted on NQ and PopQA: both use the models and memories of HotpotQA.}
\label{tab:stats}
\end{table*}

Table~\ref{tab:stats} describes the five test subsets. The training questions come from a sample of 2,000 questions per multi-hop training set (seed 42; HotpotQA restricted to \emph{hard} questions, stratified by hop count or question type). The first 1,500 fit the reliability model and fill the plan memories; the other 500 are held out. dev-A and dev-B have 500 questions each and come from the same training sets.

%======================================================================
\section{Prompts}
\label{app:prompts}

All prompts are generated from the code. The final reader uses HippoRAG~2's reading prompt and one-shot example without change.

The comparison additions are used on every dataset except MuSiQue.
\promptheading{Plan: system prompt}
\begin{lstlisting}
You split a multi-hop question into the shortest sequence of single-hop questions. Each sub-question must be answerable from one Wikipedia paragraph and ask for exactly one entity, date or number. Refer to the answer of an earlier sub-question as #1, #2, and so on. Keep every name and constraint of the original question. Do not add facts from your own knowledge and do not answer. The answer of the last sub-question must be the answer of the original question. Use as few sub-questions as possible (usually 2 to 4) and never repeat the whole original question as a sub-question.
\end{lstlisting}
\promptheading{Plan: addition for comparison questions}
\begin{lstlisting}
 Exception: for comparison or yes/no questions about two or more things, ask one sub-question for each fact that has to be compared (for example the birth date of each person) and do not ask the comparison itself; the comparison is made afterwards.
\end{lstlisting}
\promptheading{Plan: fixed examples of the single execution}
\begin{lstlisting}
Who is the spouse of the Green performer?
  -> {"hops": ["Who is the performer of Green?", "Who is the spouse of #1?"]}
When was the last time the team that Derek Jeter played for won the World Series?
  -> {"hops": ["Which team did Derek Jeter play for?", "When was the last time #1 won the World Series?"]}
In what year was the university attended by the author of The Hobbit founded?
  -> {"hops": ["Who is the author of The Hobbit?", "Which university did #1 attend?", "In what year was #2 founded?"]}
What is the capital of the county that shares a border with the county where Lake Placid is located?
  -> {"hops": ["In which county is Lake Placid located?", "Which county shares a border with #1?", "What is the capital of #2?"]}
What is the population of the city where the headquarters of the company that makes the iPhone is located?
  -> {"hops": ["Which company makes the iPhone?", "In which city is the headquarters of #1 located?", "What is the population of #2?"]}
When did the country where Niels Bohr was born join the alliance whose headquarters are in the city where the Atomium is located?
  -> {"hops": ["In which country was Niels Bohr born?", "In which city is the Atomium located?", "Which alliance has its headquarters in #2?", "When did #1 join #3?"]}
\end{lstlisting}
\promptheading{Plan: additional fixed examples for comparison questions}
\begin{lstlisting}
Who was born first, Albert Einstein or Niels Bohr?
  -> {"hops": ["When was Albert Einstein born?", "When was Niels Bohr born?"]}
Which film has the director who died later, Jaws or Psycho?
  -> {"hops": ["Who directed Jaws?", "Who directed Psycho?", "When did #1 die?", "When did #2 die?"]}
Who is the paternal grandfather of Charles III?
  -> {"hops": ["Who is the father of Charles III?", "Who is the father of #1?"]}
Are Lake Placid and Lake Tahoe both located in the same country?
  -> {"hops": ["In which country is Lake Placid located?", "In which country is Lake Tahoe located?"]}
\end{lstlisting}
\promptheading{Plan with memory: the fixed examples are replaced by the nearest stored questions and their plans}
\begin{lstlisting}
memory plan 1: the chain's own plans of training questions whose final answer was right
memory plan 2: the same, built from the plans of round 1
\end{lstlisting}
\promptheading{Hop: system prompt}
\begin{lstlisting}
You are given one question and numbered Wikipedia passages. Answer using only the passages. Return JSON with four fields in this order. passage: the number of the passage that best answers the question, or 0 if every passage is unrelated. quote: one sentence copied exactly from that passage that supports the answer, or an empty string. support: "explicit" if the quote states the answer directly; "implied" if no passage states it directly but the passage strongly suggests it (for example a place where the person worked, lived or performed, or the larger region that contains a place); "none" if the passages are unrelated. answer: the shortest exact answer (a name, place, date or number, no explanation), or an empty string when support is none. If passages are about different things with similar names, choose the one that matches every detail of the question.
\end{lstlisting}
\promptheading{Hop: second, lenient try when the first finds nothing}
\begin{lstlisting}
You are given one question and numbered Wikipedia passages. The passages may not state the answer word for word. Pick the passage most closely related to the question and give the most likely answer that it suggests (for example a place where the person lived, worked or performed, or an event named in the passage). Return JSON with four fields in this order. passage: the number of that passage, or 0 only if every passage is about something else. quote: one sentence copied exactly from that passage. support: "explicit" if the quote states the answer directly, otherwise "implied"; "none" only with passage 0. answer: the shortest answer (a name, place, date or number, no explanation).
\end{lstlisting}
\promptheading{Hop: user message}
\begin{lstlisting}
Question: <sub-question with earlier answers filled in>

Passages:
[1] <passage>

[2] <passage>
...
\end{lstlisting}
\promptheading{Hop with several candidates (variant, not part of the method)}
\begin{lstlisting}
You are given one question and numbered Wikipedia passages. Answer using only the passages. Return JSON with one field, candidates: a list of up to 3 different possible answers, the best one first. Each candidate has four fields in this order. passage: the number of the passage that supports it. quote: one sentence copied exactly from that passage. support: "explicit" if the quote states the answer directly, or "implied" if the passage only suggests it (for example a place where the person lived, worked or performed, or the larger region that contains a place). answer: the shortest exact answer (a name, place, date or number, no explanation). Add a second or third candidate only when it is a really different plausible answer, for example from a passage about another thing with a similar name, or another reading of the question. Prefer explicit candidates, but when nothing is stated directly give the most likely implied answer. Return an empty list only if every passage is unrelated to the question.
\end{lstlisting}
\promptheading{Final reader: the reading prompt of HippoRAG 2, unchanged}
\begin{lstlisting}
[system]
As an advanced reading comprehension assistant, your task is to analyze text passages and corresponding questions meticulously. Your response start after "Thought: ", where you will methodically break down the reasoning process, illustrating how you arrive at conclusions. Conclude with "Answer: " to present a concise, definitive response, devoid of additional elaborations.

[user]
Wikipedia Title: The Last Horse
The Last Horse (Spanish:El último caballo) is a 1950 Spanish comedy film directed by Edgar Neville starring Fernando Fernán Gómez.
Wikipedia Title: Southampton
The University of Southampton, which was founded in 1862 and received its Royal Charter as a university in 1952, has over 22,000 students. The university is ranked in the top 100 research universities in the world in the Academic Ranking of World Universities 2010. In 2010, the THES - QS World University Rankings positioned the University of Southampton in the top 80 universities in the world. The university considers itself one of the top 5 research universities in the UK. The university has a global reputation for research into engineering sciences, oceanography, chemistry, cancer sciences, sound and vibration research, computer science and electronics, optoelectronics and textile conservation at the Textile Conservation Centre (which is due to close in October 2009.) It is also home to the National Oceanography Centre, Southampton (NOCS), the focus of Natural Environment Research Council-funded marine research.
Wikipedia Title: Stanton Township, Champaign County, Illinois
Stanton Township is a township in Champaign County, Illinois, USA. As of the 2010 census, its population was 505 and it contained 202 housing units.
Wikipedia Title: Neville A. Stanton
Neville A. Stanton is a British Professor of Human Factors and Ergonomics at the University of Southampton. Prof Stanton is a Chartered Engineer (C.Eng), Chartered Psychologist (C.Psychol) and Chartered Ergonomist (C.ErgHF). He has written and edited over a forty books and over three hundered peer-reviewed journal papers on applications of the subject. Stanton is a Fellow of the British Psychological Society, a Fellow of The Institute of Ergonomics and Human Factors and a member of the Institution of Engineering and Technology. He has been published in academic journals including "Nature". He has also helped organisations design new human-machine interfaces, such as the Adaptive Cruise Control system for Jaguar Cars.
Wikipedia Title: Finding Nemo
Finding Nemo Theatrical release poster Directed by Andrew Stanton Produced by Graham Walters Screenplay by Andrew Stanton Bob Peterson David Reynolds Story by Andrew Stanton Starring Albert Brooks Ellen DeGeneres Alexander Gould Willem Dafoe Music by Thomas Newman Cinematography Sharon Calahan Jeremy Lasky Edited by David Ian Salter Production company Walt Disney Pictures Pixar Animation Studios Distributed by Buena Vista Pictures Distribution Release date May 30, 2003 (2003 - 05 - 30) Running time 100 minutes Country United States Language English Budget $94 million Box office $940.3 million

Question: When was Neville A. Stanton's employer founded?
Thought: 

[assistant]
The employer of Neville A. Stanton is University of Southampton. The University of Southampton was founded in 1862. 
Answer: 1862.

[user]
Wikipedia Title: <passage 1>

Wikipedia Title: <passage 2>

Wikipedia Title: <reasoning notes>

Question: <question>
Thought: 
\end{lstlisting}
\promptheading{Final reader: the reasoning notes appended after the passages}
\begin{lstlisting}
Reasoning notes checked against the passages above
Step 1: <sub-question> -> <answer>
Step 2: <sub-question> -> <answer> (likely, not stated directly)
\end{lstlisting}
\promptheading{Reader picks: system prompt}
\begin{lstlisting}
Several attempts answered the same question and they disagree. Each candidate answer is listed with the attempts that reached it. An attempt either worked step by step (each step shows its sub-question, its answer and the sentence it quoted from a Wikipedia passage) or read the top retrieved passages in one go (no steps to show). Decide which candidate answer is best supported: the steps must follow the question, every step must be backed by its quote, and a step marked "likely", a missing quote or an attempt marked "incomplete" is weaker evidence. Return JSON with two fields in this order. reason: one or two sentences. choice: the number of the best candidate.
\end{lstlisting}

%======================================================================
\section{Implementation Details}
\label{app:impl}

\begin{table}[!htbp]
\centering
\footnotesize
\begin{tabular}{@{}p{0.38\columnwidth}p{0.56\columnwidth}@{}}
\toprule
Item & Setting \\
\midrule
Reader and planner & Qwen3.5-9B, thinking off, served by vLLM \\
Embedder & NV-Embed-v2 \\
Decoding & temperature 0, top-$p$ 1.0, seed 42; samples at $T{=}0.7$ or $0.3$ \\
Max new tokens & plan 400, hop 300, final reader 2,048 \\
Passages per hop & 8 \\
Passages for the final reader & 5, with the hop notes \\
Answers per hop & 1 \\
Plan examples & 6 fixed (plus 4 for comparison questions), or the 6 nearest training questions in memory \\
Reliability model & logistic regression, L2 penalty 1.0, 1,500 training questions, label F1 $\geq 0.5$ \\
Answer groups & token-level match after normalisation \\
On demand & plan executions run only when the dense read and the single execution disagree \\
Intervals & paired bootstrap over questions, 10,000 resamples, seed 0 \\
Hardware & one GPU, 12 questions in parallel per run \\
\bottomrule
\end{tabular}
\caption{Implementation settings.}
\label{tab:impl}
\end{table}

Table~\ref{tab:impl} lists the settings. Every LLM call is cached, so running a command again reproduces its outputs. Scoring uses HippoRAG~2's answer extraction and EM/F1 functions.

%======================================================================
\section{Reliability Features and Weights}
\label{app:features}

\begin{table}[!htbp]
\centering
\scriptsize
\setlength{\tabcolsep}{3pt}
\begin{tabular}{@{}lp{0.62\columnwidth}@{}}
\toprule
Feature & Definition \\
\midrule
\multicolumn{2}{@{}l}{\textit{Completion and agreement}} \\
\texttt{complete} & the chain bound an answer at every hop \\
\texttt{chain\_reader\_agree} & the last hop's answer matches the final reader's answer \\
\texttt{marker} & the final reader's output has a marked answer \\
\multicolumn{2}{@{}l}{\textit{Quote evidence}} \\
\texttt{min\_explicit} & every hop is marked explicit (minimum over hops) \\
\texttt{mean\_verbatim} & share of hops whose quote occurs verbatim in the cited passage \\
\texttt{min\_answer\_in\_quote} & every hop's answer occurs in its quote \\
\multicolumn{2}{@{}l}{\textit{Retrieval scores}} \\
\texttt{max\_rank} & worst retrieval rank of a cited passage \\
\texttt{min\_passage\_score} & lowest retrieval score of a cited passage \\
\texttt{min\_question\_score} & lowest score between a sub-question and its passages \\
\texttt{min\_next\_top1} & top-1 retrieval score of the following hop (minimum) \\
\multicolumn{2}{@{}l}{\textit{Plan shape}} \\
\texttt{hops} & number of hops (scaled) \\
\texttt{linked} & the hops are linked through earlier answers (\#1, \#2) \\
\texttt{coverage} & how much of the question the plan covers \\
\texttt{answer\_in\_question} & the answer occurs in the question \\
\texttt{backtracked} & the chain backtracked (always 0 in the method) \\
\texttt{one\_hop} & the plan has a single hop \\
\bottomrule
\end{tabular}
\caption{Features of the reliability model, which also has a bias term. Features are standardised with the mean and scale of the training executions.}
\label{tab:features}
\end{table}

\begin{figure*}[!htbp]
\centering
\includegraphics[width=\textwidth]{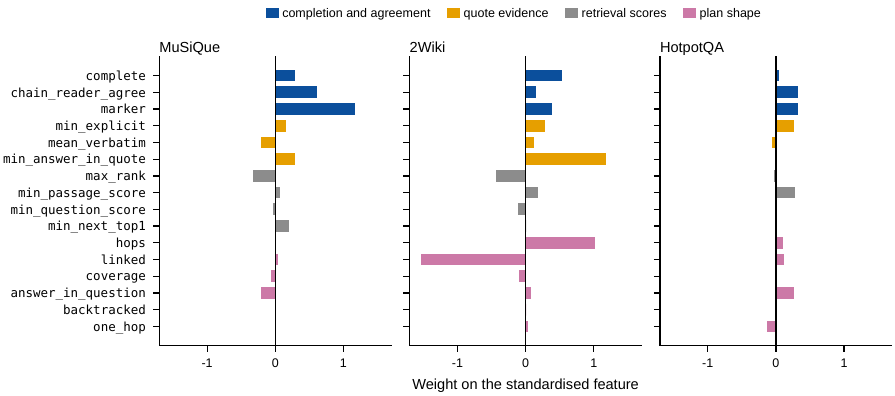}
\caption{Weights of the reliability models on standardised features, one model per multi-hop dataset; NQ and PopQA use the HotpotQA model. Intercepts: 0.45 MuSiQue, 3.37 2Wiki, 1.70 HotpotQA. The dense read gets a constant reliability, its share of right answers on the training questions: 0.46, 0.60 and 0.78.}
\label{fig:weights}
\end{figure*}

Table~\ref{tab:features} defines the features and Figure~\ref{fig:weights} shows the fitted weights. Refitting on the same training questions reproduces these weights exactly.

%======================================================================
\section{Baselines}
\label{app:baselines}

All same-reader baselines use Qwen3.5-9B and NV-Embed-v2. The dense read gives the reader the top five passages with HippoRAG~2's reading prompt. HippoRAG~2 runs from its official code, with our reader for extraction and reading.

\paragraph{Replication check.} With our reader, HippoRAG~2 reaches 45.8, 72.3, 74.7, 58.4 and 55.8 F1; the original paper reports 48.6, 71.0, 75.5, 63.3 and 56.2 with Llama-3.3-70B. The difference is within 3 points on four datasets; on NQ ours is 4.9 lower, and the same-reader dense read on NQ is also 3.9 below the original paper's dense retrieval (58.0 against 61.9), so the gap follows the reader.

\paragraph{Extraction on NQ and PopQA.} The official code stops a whole corpus when extraction fails to parse for any passage. One NQ passage (of 9,633) and two PopQA passages (of 8,676) failed on every attempt. For the two PopQA passages, the model wrote each entity as an object with a type label; our wrapper takes out the entity names, so nothing is lost. For the NQ passage, the model returned a JSON object without an entity list; we continue with no entities, which is what the official single-passage function returns, and allow at most five such passages per corpus. The changes are in our wrapper only, not in the official code, and are logged. The three multi-hop corpora never hit either case. After the fix, extraction, encoding, graph building and answering ran with zero errors; extraction used 14.0M tokens on NQ and 13.4M on PopQA.

%======================================================================
\section{Retrieval Quality and Quotes}
\label{app:retrieval}

\begin{table*}[!htbp]
\centering
\footnotesize
\begin{tabular}{@{}lccccc@{}}
\toprule
Passages in front of the reader & MuSiQue & 2Wiki & HotpotQA & NQ & PopQA \\
\midrule
Dense read: top 5 & 69.3 / 37.4 & 76.6 / 50.5 & 94.4 / 89.1 & 75.0 / 43.5 & 51.0 / 2.2 \\
IRCoT: top 5 & 71.6 / 40.7 & 79.9 / 57.7 & 95.5 / 91.2 & -- & -- \\
HippoRAG 2: top 5 & 74.0 / -- & 92.7 / -- & 95.7 / -- & 76.3 / -- & 51.9 / -- \\
Single execution: final reader's 5 & 81.6 / 58.8 & 97.8 / 94.8 & 97.3 / 94.8 & 74.4 / 42.7 & 51.1 / 2.4 \\
Plan 1: final reader's 5 & 82.2 / 60.2 & 97.9 / 95.0 & 97.5 / 95.3 & 74.9 / 43.4 & 51.2 / 2.6 \\
Plan 2: final reader's 5 & 81.8 / 59.1 & 97.9 / 95.0 & 97.5 / 95.1 & 74.6 / 43.1 & 51.2 / 2.6 \\
\bottomrule
\end{tabular}
\caption{Retrieval quality on test: recall of gold evidence passages among the five passages given to the reader / share of questions with all gold passages present (\%). HippoRAG 2 reports recall only.}
\label{tab:retrieval}
\end{table*}

\begin{table}[!htbp]
\centering
\scriptsize
\setlength{\tabcolsep}{2pt}
\begin{tabular}{@{}lrrrrrr@{}}
\toprule
Dataset & Hops & \shortstack[r]{Verbatim\\(\%)} & \shortstack[r]{Explicit\\(\%)} & \shortstack[r]{Complete, all\\verbatim (\%)} & \shortstack[r]{F1 of\\those} & \shortstack[r]{F1 of\\the rest} \\
\midrule
MuSiQue & 2,564 & 95.6 & 86.0 & 88.6 & 64.4 & 23.8 \\
2Wiki & 2,488 & 99.0 & 94.4 & 97.4 & 78.8 & 55.2 \\
HotpotQA & 2,013 & 97.8 & 94.5 & 95.6 & 76.9 & 67.2 \\
NQ & 1,403 & 97.2 & 90.4 & 96.1 & 60.4 & 44.2 \\
PopQA & 1,153 & 95.1 & 85.7 & 94.3 & 60.1 & 11.9 \\
\bottomrule
\end{tabular}
\caption{Quotes of the single execution on test. A quote is verbatim when it occurs word for word in the retrieved passage it cites. The last three columns split the questions into those whose chain finished with every quote verbatim and the rest.}
\label{tab:quotes}
\end{table}

\begin{table*}[!tbp]
\begin{minipage}[t]{\dimexpr(\textwidth-\columnsep)/2\relax}
\vspace{0pt}
\captionsetup{type=table}
\centering
\footnotesize
\setlength{\tabcolsep}{3pt}
\begin{tabular}{@{}lrrr@{}}
\toprule
MuSiQue split & Right & Missing & Not bound \\
\midrule
dev-A (500) & 61.6 & 5.2 & 33.2 \\
Held-out training (500) & 64.8 & 6.8 & 28.4 \\
\bottomrule
\end{tabular}
\caption{Root cause of wrong single executions on MuSiQue development questions (\% of questions): the first hop of the gold decomposition that was not bound to the right answer, split by whether its supporting passage was missing from retrieval or retrieved but not bound.}
\label{tab:rootcause}
\end{minipage}\hfill
\begin{minipage}[t]{\dimexpr(\textwidth-\columnsep)/2\relax}
\vspace{0pt}
\captionsetup{type=table}
\centering
\footnotesize
\begin{tabular}{@{}p{0.78\columnwidth}r@{}}
\toprule
Binding error (dev-A, 166 questions) & Count \\
\midrule
Another answer from the supporting passage & 65 \\
Answer taken from another passage & 48 \\
Plan has a different number of hops than the gold decomposition & 32 \\
No answer bound & 14 \\
Supporting passage used only at another hop & 7 \\
\bottomrule
\end{tabular}
\caption{The 166 dev-A questions of Table~\ref{tab:rootcause} whose supporting passage was retrieved but not bound. Most ``another answer'' cases differ in granularity or wording.}
\label{tab:binding}
\end{minipage}
\end{table*}

Table~\ref{tab:retrieval} gives evidence recall, Table~\ref{tab:quotes} the quote statistics, and Tables~\ref{tab:rootcause} and~\ref{tab:binding} the root causes of wrong answers on MuSiQue development questions. The method never uses gold decompositions; apart from this analysis, they are used only by the two chain variants marked $^\ddagger$ in Table~\ref{tab:chain}.

%======================================================================
\section{Sources of Executions and All Merge Comparisons}
\label{app:sources}
\label{app:merge}

\paragraph{Merge rules.} Answers are grouped by token-level match: after normalisation, an answer joins a group when its tokens are contained in, or contain, those of the group's first answer. Executions are taken in the order dense read, single execution, plan~1, plan~2 (then samples).
\emph{Support} returns the answer of the most reliable execution in the group with the highest support.
\emph{Majority vote} counts the executions in each group and returns the answer of the first member of the largest group; the first member is the dense read whenever the dense read is in that group.
\emph{Most reliable execution} returns the answer of the execution with the highest reliability, without pooling.
\emph{Reader picks} calls the reader with the prompt of Appendix~\ref{app:prompts} only when the executions disagree, which adds 0.16 to 0.59 calls per question; on dev-A, support is higher by 2.9, 2.0 and $-$0.1 F1 on the three multi-hop datasets.

\begin{table*}[!htbp]
\centering
\footnotesize
\begin{tabular}{@{}llllllr@{}}
\toprule
Executions merged by support & MuSiQue & 2Wiki & HotpotQA & NQ & PopQA & Calls/q \\
\midrule
Single execution & 59.8 & 78.2 & 76.5 & 59.8 & 57.4 & 5.3 \\
+ 2 plans & 62.7\diff{+2.9}$^\dagger$ & 79.4\diff{+1.2}$^\dagger$ & 77.2\diff{+0.8} & 61.6\diff{+1.9}$^\dagger$ & 57.1\diff{$-$0.3} & 15.9 \\
+ 2 samples, $T{=}0.7$ & 62.6\diff{+2.8}$^\dagger$ & 79.5\diff{+1.3}$^\dagger$ & 78.0\diff{+1.5}$^\dagger$ & 61.4\diff{+1.6}$^\dagger$ & 59.0\diff{+1.6}$^\dagger$ & 16.3 \\
+ 2 samples, $T{=}0.3$ & 64.2\diff{+4.4}$^\dagger$ & 79.1\diff{+1.0}$^\dagger$ & 77.6\diff{+1.2}$^\dagger$ & -- & -- & 16.1 \\
+ 3 samples, $T{=}0.7$ & 64.1\diff{+4.3}$^\dagger$ & 79.3\diff{+1.1}$^\dagger$ & 79.1\diff{+2.6}$^\dagger$ & -- & -- & 21.7 \\
\midrule
Dense read + single + 2 plans (full) & 63.0\diff{+3.2}$^\dagger$ & 78.6\diff{+0.5} & 78.0\diff{+1.5}$^\dagger$ & 61.4\diff{+1.7}$^\dagger$ & 57.4\diff{0.0} & 16.9 \\
Dense read + single + 2 samples, $T{=}0.7$ & 62.6\diff{+2.8}$^\dagger$ & 79.0\diff{+0.8} & 78.8\diff{+2.4}$^\dagger$ & 61.9\diff{+2.2}$^\dagger$ & 58.1\diff{+0.7} & 17.3 \\
Dense read + single + 2 samples, $T{=}0.3$ & 63.6\diff{+3.8}$^\dagger$ & 78.5\diff{+0.4} & 77.9\diff{+1.4}$^\dagger$ & -- & -- & 17.1 \\
Dense read + single + 3 samples, $T{=}0.7$ & 64.3\diff{+4.5}$^\dagger$ & 79.1\diff{+1.0} & 79.0\diff{+2.5}$^\dagger$ & -- & -- & 22.7 \\
All nine executions & 64.3\diff{+4.5}$^\dagger$ & 80.1\diff{+1.9}$^\dagger$ & 78.9\diff{+2.4}$^\dagger$ & -- & -- & 44.0 \\
\bottomrule
\end{tabular}
\caption{Where the extra executions come from (test F1). Small numbers: minus the single execution, paired; $^\dagger$: the 95\% interval excludes zero. Plans: the chain with plan examples taken from the chain's own right plans on training questions. Samples: the chain decoded again at temperature $T$. Calls per question are for MuSiQue.}
\label{tab:source}
\end{table*}

\begin{table}[!htbp]
\centering
\footnotesize
\setlength{\tabcolsep}{1.8pt}
\begin{tabular}{@{}lrrrrr@{}}
\toprule
 & MuSiQue & 2Wiki & Hotpot & NQ & PopQA \\
\midrule
\multicolumn{6}{@{}l}{\textit{Full method}} \\
\quad Support F1 & 63.0 & 78.6 & 78.0 & 61.4 & 57.4 \\
\quad $-$ vote & +3.42$^\dagger$ & +1.76$^\dagger$ & $-$0.33 & +2.23$^\dagger$ & $-$0.04 \\
\quad $-$ most rel. & +0.29 & +0.41 & +0.05 & +1.05 & +1.87$^\dagger$ \\
\multicolumn{6}{@{}l}{\textit{Single + 2 plans}} \\
\quad Support F1 & 62.7 & 79.4 & 77.2 & 61.6 & 57.1 \\
\quad $-$ vote & +1.74$^\dagger$ & +0.60$^\dagger$ & +0.04 & +0.71 & $-$0.02 \\
\quad $-$ most rel. & +0.25 & $-$0.20 & $-$0.04 & +1.00 & $-$0.16 \\
\multicolumn{6}{@{}l}{\textit{Single + 2 samples, $T{=}0.7$}} \\
\quad Support F1 & 62.6 & 79.5 & 78.0 & 61.4 & 59.0 \\
\quad $-$ vote & +1.77$^\dagger$ & +1.05$^\dagger$ & +0.81$^\dagger$ & +0.09 & +1.02$^\dagger$ \\
\quad $-$ most rel. & +0.66 & $-$0.14 & +0.28 & +1.55$^\dagger$ & $-$0.17 \\
\multicolumn{6}{@{}l}{\textit{Single + 3 samples, $T{=}0.7$}} \\
\quad Support F1 & 64.1 & 79.3 & 79.1 & -- & -- \\
\quad $-$ vote & +2.34$^\dagger$ & +0.79 & +1.26$^\dagger$ & -- & -- \\
\quad $-$ most rel. & +1.50$^\dagger$ & +0.01 & +1.26$^\dagger$ & -- & -- \\
\multicolumn{6}{@{}l}{\textit{Single + 2 samples, $T{=}0.3$}} \\
\quad Support F1 & 64.2 & 79.1 & 77.6 & -- & -- \\
\quad $-$ vote & +2.49$^\dagger$ & +0.56$^\dagger$ & +0.36 & -- & -- \\
\quad $-$ most rel. & +1.13$^\dagger$ & +0.03 & +0.63 & -- & -- \\
\bottomrule
\end{tabular}
\caption{Support against majority vote and the most reliable execution for each set of executions (test): support F1, and support minus majority vote or minus the most reliable execution (\emph{most rel.}; paired). $^\dagger$: the 95\% interval excludes zero.}
\label{tab:mergeall}
\end{table}

Table~\ref{tab:source} compares the sources of extra executions. The single execution plus two samples at $T{=}0.7$ differs from the full method, which uses two plan executions from memory and the dense read, by $-$0.40, +0.88, 0.00 and $-$0.03 F1 on the first four datasets (all intervals include zero) and by +1.54 on PopQA, in favour of sampling. Table~\ref{tab:mergeall} compares support with majority voting and with the most reliable execution on further sets of executions; support is never significantly below either.

%======================================================================
% Queue the wide exhibits together above the discussions in I and J.
\begin{table*}[!htbp]
\centering
\footnotesize
\setlength{\tabcolsep}{6pt}
\begin{tabular}{@{}lrrrrr@{}}
\toprule
 & MuSiQue & 2Wiki & HotpotQA & NQ & PopQA \\
\midrule
All four executions agree (\%) & 40.1 & 62.0 & 81.8 & 62.1 & 83.9 \\
Full method F1 when they agree & 81.1 & 86.9 & 83.7 & 75.6 & 63.0 \\
Full method F1 when they differ & 50.9 & 65.2 & 52.4 & 38.1 & 28.5 \\
Single execution F1 when they differ & 46.2 & 64.2 & 44.0 & 32.8 & 28.7 \\
Dense read and single execution agree (\%) & 44.9 & 63.4 & 84.6 & 68.7 & 84.9 \\
\bottomrule
\end{tabular}
\caption{Agreement among the four executions of the full method (test).}
\label{tab:agree}
\end{table*}

\begin{figure*}[!htbp]
\centering
\includegraphics[width=\textwidth]{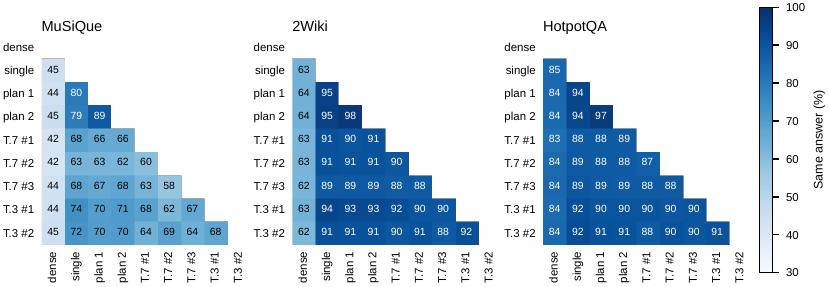}
\caption{Share of test questions on which two executions give the same answer. single: single execution; plan 1, plan 2: plan examples from memory; T.7, T.3: samples at temperature 0.7 and 0.3.}
\label{fig:agreement}
\end{figure*}

\begin{table*}[!htbp]
\centering
\footnotesize
\begin{tabular}{@{}llccccc@{}}
\toprule
 & & \multicolumn{5}{c}{Cosine similarity of the question to its nearest training question} \\
\cmidrule(l){3-7}
Dataset & Split & $<0.3$ & 0.3--0.4 & 0.4--0.5 & 0.5--0.6 & $\geq 0.6$ \\
\midrule
MuSiQue & test & 555\,(+0.8) & 360\,(+0.1) & 76\,($-$3.6) & 8 & 1 \\
2Wiki & test & 155\,(0.0) & 375\,(+0.7) & 334\,(+0.2) & 112\,(+1.6) & 24\,(+4.2) \\
HotpotQA & test & 704\,(+0.9) & 248\,($-$1.0) & 33\,(+5.1) & 12 & 3 \\
\midrule
MuSiQue & dev-A & 58\,($-$1.6) & 93\,($-$0.3) & 68\,(+3.7) & 103\,(+6.1) & 178\,(+5.9) \\
2Wiki & dev-A & 66\,($-$1.5) & 154\,($-$0.7) & 177\,($-$1.1) & 83\,(+1.6) & 20\,(+5.0) \\
HotpotQA & dev-A & 347\,(+0.3) & 126\,(+1.0) & 19 & 5 & 3 \\
\bottomrule
\end{tabular}
\caption{Plan examples from memory against the fixed examples, by how close the question is to the training questions in memory: number of questions (F1 of plan 2 minus F1 of the single execution; omitted for bins under 20 questions).}
\label{tab:similarity}
\end{table*}

\FloatBarrier
\section{Agreement Between Executions}
\label{app:agreement}

When all four executions agree (40.1\% of MuSiQue questions), the full method reaches 81.1 F1; when they differ, it reaches 50.9 against 46.2 for the single execution (Table~\ref{tab:agree}). The two plan executions from memory agree with each other on 88.8\% of MuSiQue questions, more than any two samples (57.8\% to 68.8\%), so samples add more diversity per execution (Figure~\ref{fig:agreement}).

%======================================================================
% Begin J in the right column beside I.
\newpage
\section{Plan Examples from Memory and Other Chain Variants}
\label{app:variants}

Plan examples from memory work by finding a similar training question. On test, most questions are far from every training question (Table~\ref{tab:similarity}), and on MuSiQue memory plans are not significantly better than the fixed examples in any similarity bin with at least 20 questions. On MuSiQue dev-A, where questions are closer, memory plans gain 5.9 to 6.1 F1 above similarity 0.5.
On MuSiQue dev-B, the learned hop scorer with backtracking is 3.79 F1 below the single execution. Plan examples from gold decompositions need decomposition annotations, which the method does not use.

%======================================================================
% The cost breakdown, plots and case transcripts use the full text width.
\onecolumn
\ifdefined\leftlinenumbers\leftlinenumbers\fi
\section{Cost and Running Time}
\label{app:cost}

\begin{figure*}[!htbp]
\centering
\includegraphics[width=\textwidth]{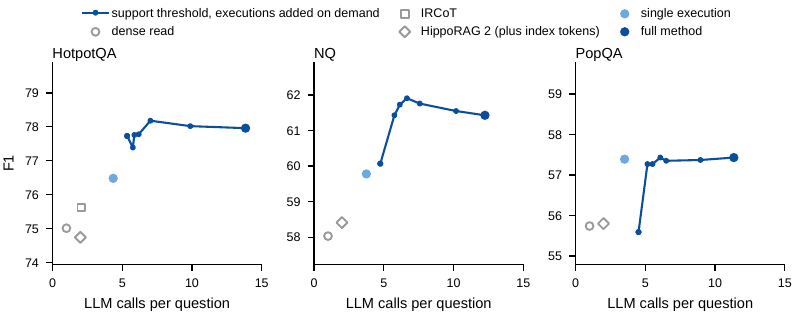}
\caption{Test F1 against LLM calls per question on HotpotQA, NQ and PopQA, as in Figure~\ref{fig:cost}. HippoRAG~2 also spends 14.1M, 14.0M and 13.4M index tokens on these corpora.}
\label{fig:costother}
\end{figure*}

Table~\ref{tab:cost} gives the full cost, Figure~\ref{fig:costother} the cost--accuracy curves of the other datasets, and Table~\ref{tab:wallclock} the running time.
The on-demand rule saves 26\% to 51\% of the calls of the full method; its F1 differs from the full method's significantly only on 2Wiki ($-$0.40).

\begin{table*}[!htbp]
\centering
\footnotesize
\begin{tabular}{@{}llrrrrr@{}}
\toprule
Dataset & System & Index tokens & Preparation (once) & Calls/q & Tokens/q & Index + 1,000 q \\
\midrule
MuSiQue & Dense read & 0 & 0 & 1.0 & 2,129 & 2.1M \\
 & Single execution & 0 & 0 & 5.3 & 7,089 & 7.1M \\
 & Full method & 0 & 33.5M & 16.9 & 23,171 & 23.2M \\
 & Full method, on demand & 0 & 33.5M & 12.5 & 17,535 & 17.5M \\
 & IRCoT & 0 & 0 & 2.3 & 4,389 & 4.4M \\
 & HippoRAG 2 & 15.6M & 0 & 2.0 & 5,170 & 20.8M \\
\midrule
2Wiki & Dense read & 0 & 0 & 1.0 & 1,996 & 2.0M \\
 & Single execution & 0 & 0 & 4.8 & 6,799 & 6.8M \\
 & Full method & 0 & 21.2M & 15.3 & 21,710 & 21.7M \\
 & Full method, on demand & 0 & 21.2M & 9.4 & 13,719 & 13.7M \\
 & IRCoT & 0 & 0 & 2.2 & 3,611 & 3.6M \\
 & HippoRAG 2 & 8.3M & 0 & 2.0 & 5,069 & 13.4M \\
\midrule
HotpotQA & Dense read & 0 & 0 & 1.0 & 1,607 & 1.6M \\
 & Single execution & 0 & 0 & 4.4 & 5,616 & 5.6M \\
 & Full method & 0 & 18.5M & 13.8 & 17,783 & 17.8M \\
 & Full method, on demand & 0 & 18.5M & 6.7 & 8,991 & 9.0M \\
 & IRCoT & 0 & 0 & 2.1 & 3,479 & 3.5M \\
 & HippoRAG 2 & 14.1M & 0 & 2.0 & 4,698 & 18.8M \\
\midrule
NQ & Dense read & 0 & 0 & 1.0 & 1,744 & 1.7M \\
 & Single execution & 0 & 0 & 3.8 & 5,081 & 5.1M \\
 & Full method & 0 & 18.5M & 12.3 & 16,672 & 16.7M \\
 & Full method, on demand & 0 & 18.5M & 7.2 & 10,142 & 10.1M \\
 & HippoRAG 2 & 14.0M & 0 & 2.0 & 4,809 & 18.9M \\
\midrule
PopQA & Dense read & 0 & 0 & 1.0 & 1,643 & 1.6M \\
 & Single execution & 0 & 0 & 3.5 & 4,584 & 4.6M \\
 & Full method & 0 & 18.5M & 11.3 & 14,740 & 14.7M \\
 & Full method, on demand & 0 & 18.5M & 5.7 & 7,802 & 7.8M \\
 & HippoRAG 2 & 13.4M & 0 & 2.0 & 4,698 & 18.1M \\
\bottomrule
\end{tabular}
\caption{LLM cost on test (prompt plus completion tokens). \emph{Index}: LLM calls over the corpus before any question. \emph{Preparation}: runs on the 1,500 training questions that fit the reliability model and fill the plan memories, once per dataset; NQ and PopQA reuse those of HotpotQA. The last column adds the index to 1,000 questions.}
\label{tab:cost}
\end{table*}

\begin{table}[!htbp]
\centering
\footnotesize
\setlength{\tabcolsep}{2.4pt}
\begin{tabular}{@{}lrrrrr@{}}
\toprule
Minutes & MuSiQue & 2Wiki & Hotpot & NQ & PopQA \\
\midrule
Dense read & 40.5 & 16.3 & 8.3 & 9.7 & 7.7 \\
Single exec. & 27.4 & 22.7 & 19.4 & 17.0 & 13.4 \\
Plan 1 & 16.0 & 8.3 & 11.3 & 11.7 & 4.0 \\
Plan 2 & 10.0 & 5.3 & 7.7 & 7.3 & 2.3 \\
Sample, $T{=}0.7$ & 32.0 & 23.7 & 21.7 & 20.7 & 16.0 \\
IRCoT & 16.6 & 22.4 & 14.3 & -- & -- \\
HR2 extraction & 119.5 & 66.4 & 112.4 & 85.3 & 87.2 \\
HR2 encoding & 43.0 & 22.7 & 42.0 & 38.7 & 39.3 \\
HR2 graph & 5.7 & 2.0 & 5.3 & 4.0 & 4.3 \\
\bottomrule
\end{tabular}
\caption{Wall-clock minutes per test job on one GPU, 12 questions in parallel for the chain jobs. Jobs that ran later reuse cached LLM calls of earlier jobs, which shortens them. HR2: HippoRAG~2. Answering with HippoRAG~2 ran as one job per batch of datasets: 232.7 minutes for the three multi-hop datasets and 113.7 for NQ and PopQA.}
\label{tab:wallclock}
\end{table}

%======================================================================
\FloatBarrier
\section{More Cases}
\label{app:cases}

Cases are chosen by fixed rules from the test predictions: for each outcome category and dataset, up to eight questions in decreasing order of support. Table~\ref{tab:cases} shows one case from each of three categories.

% Cases from data/cases.jsonl (test). Kinds: "support corrects A1" (2Wiki 3c18a0d00bb011ebab90acde48001122),
% "support overrides a right A1" (MuSiQue 3hop2__89854_90327_76291), "no execution right" (MuSiQue 2hop__13731_801817).
\begin{table}[!htbp]
\centering
\footnotesize
\setlength{\tabcolsep}{3pt}
\renewcommand{\arraystretch}{1.1}
\begin{tabular}{@{}p{0.985\linewidth}@{}}
\toprule
\textbf{(1) Support corrects the single execution} (2Wiki). \textbf{Q:} Who is the paternal grandmother of George Rodney, 3rd Baron Rodney? \quad \textbf{Gold:} Lady Jane Compton \\
\textbf{Single execution}, $r=0.92$: 1.~Who is the father of George Rodney, 3rd Baron Rodney? $\rightarrow$ George Rodney, 2nd Baron Rodney; 2.~Who is the \emph{wife} of George Rodney, 2nd Baron Rodney? $\rightarrow$ Anne Harley. \hfill F1 0.0 \\
\textbf{Plan executions from memory} (two), $r=0.93$ each: 1.~as above; 2.~Who is the \emph{mother} of George Rodney, 2nd Baron Rodney? $\rightarrow$ Lady Jane Compton, quoting \textit{``Rodney was the son of Admiral George Rodney, 1st Baron Rodney, by his first wife Lady Jane Compton \dots''} \hfill F1 1.0 \\
\textbf{Dense read}, $r=0.60$: Lady Jane Compton. \quad \textbf{Support:} Lady Jane Compton, $S=0.998$ (F1 1.0). \quad The single execution plans the wrong relation; every quote is verbatim. \\
\midrule
\textbf{(2) Support overrides a right single execution} (MuSiQue). \textbf{Q:} When did the party that hold the majority in the House of Reps take control of the branch that approves members of the American cabinet? \quad \textbf{Gold:} January 2015 \\
\textbf{Single execution}, $r=0.82$: Republicans $\rightarrow$ Senate $\rightarrow$ \textbf{January 2015}, quoting \textit{``The Republicans regained the majority of the Senate in the 114th Congress, which started in January 2015 \dots''} \hfill F1 1.0 \\
\textbf{Plan executions from memory} (two), $r=0.80$ each: Senate $\rightarrow$ \textbf{2014}, quoting \textit{``The 2014 elections gave the Republicans control of the Senate \dots''} \hfill F1 0.0 \\
\textbf{Dense read}, $r=0.46$: 2014. \quad \textbf{Support:} 2014, $S=0.979$ (F1 0.0). \quad Three executions agree on the election year, which pools above the single right execution. \\
\midrule
\textbf{(3) No execution is right} (MuSiQue). \textbf{Q:} Where was Mary's betrothed born? \quad \textbf{Gold:} Nazareth \\
\textbf{All three chain executions}, $r=0.34$ each: 1.~Who is Mary's betrothed? $\rightarrow$ Joseph; 2.~Where was Joseph born? $\rightarrow$ Fairview, Kentucky, quoting \textit{``Davis was born in Fairview, Kentucky, to a moderately prosperous farmer \dots''} \hfill F1 0.0 \\
\textbf{Dense read}, $r=0.46$: ``Not mentioned in the text''. \quad \textbf{Support:} Fairview, Kentucky, $S=0.71$ (F1 0.0). \quad The second hop binds a passage about another person (Jefferson Davis); the quote is verbatim but off-topic, and the low reliability marks the execution as doubtful. \\
\bottomrule
\end{tabular}
\caption{More test cases, chosen by fixed rules from three outcome categories. Reliabilities and supports are as computed by the full method.}
\label{tab:cases}
\end{table}

\end{document}